\documentclass{article}

\usepackage[utf8]{inputenc}
\usepackage[T1]{fontenc}
\usepackage{lmodern}
\usepackage[margin=1in]{geometry}
\usepackage{amsmath,amssymb}
\usepackage{booktabs}
\usepackage{graphicx}
\usepackage{hyperref}
\usepackage{caption}
\usepackage{float}

\usepackage{titlesec}
\titleformat{\section}{\bfseries\large\MakeUppercase}{}{0em}{}
\titleformat{\subsection}{\bfseries}{}{0em}{}
\titleformat{\subsubsection}{\normalfont}{}{0em}{}
\titleformat{\paragraph}[runin]{\itshape}{}{0em}{}

\hypersetup{
    colorlinks=true,
    linkcolor=blue,
    citecolor=blue,
    urlcolor=blue
}

\date{}

\begin{document}

\begin{titlepage}
\centering
\vspace*{2cm}

{\LARGE\bfseries Case-Specific Rubrics for Clinical AI Evaluation: Methodology, Validation, and LLM-Clinician Agreement Across 823 Encounters\par}

\vspace{2cm}

Aaryan Shah$^{1,2,*}$, Andrew Hines$^{1}$, Alexia Downs$^{1}$, Denis Bajet$^{1}$, Paulius Mui, MD$^{3}$, Fabiano Araujo, MD, PhD$^{4}$, Laura Offutt, MD$^{3}$, Aida Rutledge, MD$^{5}$, Elizabeth Jimenez$^{3}$

\vspace{1cm}

$^{1}$Canvas Medical, San Francisco, CA, USA\\
$^{2}$Department of Biomedical Data Science, Stanford University, Stanford, CA, USA\\
$^{3}$XPC (X Primary Care)\\
$^{4}$FCA Consulting\\
$^{5}$Department of Pediatrics, University of Nevada, Reno, NV, USA\\

\vspace{1cm}

$^{*}$Corresponding author:\\
Aaryan Shah\\
Canvas Medical\\
San Francisco, CA, USA\\
aaryan.shah@canvasmedical.com

\vspace{1cm}


\end{titlepage}

\section*{Abstract}
                                    
\textit{\textbf{Objective.}} Clinical AI systems generating documentation from patient encounters require evaluation methodologies that are clinically valid, economically viable, and sensitive to iterative changes. Methods requiring expert review per scoring instance are too slow and expensive for safe, iterative deployment. We present a case-specific, clinician-authored rubric methodology for clinical AI evaluation and examine whether LLM-generated rubrics can approximate clinician agreement.                           
                                                                                                                                       
\textit{\textbf{Materials and Methods.}} Twenty clinicians authored 1,646 rubrics for 823 clinical cases (736 real-world, 87 synthetic) across primary care, psychiatry, oncology, and behavioral health. Each rubric was validated by confirming that an LLM-based scoring agent consistently scored clinician-preferred outputs higher than rejected ones. Seven versions of an EHR-embedded AI agent for clinicians were evaluated across all cases.

\textit{\textbf{Results.}} Clinician-authored rubrics discriminated effectively between high- and low-quality outputs (median score gap: 82.9\%) with high scoring stability (median range: 0.00\%). Median scores improved from 84\% to 95\%. In later experiments, clinician-LLM ranking agreement (tau: 0.42--0.46) matched or exceeded clinician-clinician agreement (tau: 0.38--0.43), attributable to both ceiling compression and LLM rubric improvement.

\textit{\textbf{Discussion.}} This convergence supports incorporating LLM-generated rubrics alongside clinician-authored ones. At roughly 1,000 times lower cost, LLM rubrics enable substantially greater evaluation coverage, while continued clinical authorship grounds evaluation in expert judgment. Ceiling compression poses a methodological challenge for future inter-rater agreement studies.
                                         
\textit{\textbf{Conclusion.}} Case-specific rubrics offer a path for clinical AI evaluation that preserves expert judgment while enabling automation at three orders of magnitude lower cost. Clinician-authored rubrics establish the baseline against which LLM rubrics are validated.    

\section{Introduction}
Clinical AI systems are increasingly deployed in healthcare, yet evaluation methodologies remain intermittent at best, and disconnected from real-world clinical practice \cite{andersen2024,brodeur2026}. Expert clinician judgment is widely regarded as the gold standard for assessing clinical AI outputs, but it does not scale \cite{croxford2025,tam2024}. The time, cost, and cognitive demands of physician review create an insurmountable barrier to iterative development and continuous improvement.

Correctness in clinical AI documentation is both patient-dependent and scenario-dependent. The information that constitutes an appropriate clinical note varies with the patient's history, the clinical domain, the encounter type, and the documentation conventions of the practice setting \cite{walker2017,burke2023}. Even for the same encounter, expert clinicians may disagree on what constitutes adequate documentation, reflecting differences in specialty norms, risk tolerance, and documentation philosophy \cite{sylolypavan2023}. This variability demands evaluation approaches grounded in case-specific expert judgment rather than universal scoring criteria.

Prior work has advanced clinical AI evaluation along several dimensions, including offline benchmark suites, physician-authored rubrics for conversational AI, diagnostic reasoning assessments, and prospective deployment trials \cite{arora2025,medhelm,maidxo,penda}. However, these efforts share structural limitations. Most operate without interaction with live electronic health records or longitudinal patient context. Those conducted in clinical settings remain tightly coupled to single deployment contexts with limited generalizability. None bridges the gap between controlled bench evaluation and continuous field-level validation within live electronic health record (EHR) workflows.

The core challenge is encoding expert clinical judgment in a form that reduces the cost of repeated evaluation. Physician review produces high-validity assessments but cannot be applied continuously across model iterations, prompt changes, and architectural modifications at the cadence required by modern AI development \cite{croxford2025,tam2024}. What is needed is a methodology that captures the specificity of expert evaluation \textemdash \space scenario-specific, context-dependent, clinically grounded \textemdash \space while enabling repeated application without physician involvement for each scoring instance.

Rubric-based approaches offer a potential bridge: if expert judgment can be encoded as structured, case-specific evaluation criteria, those criteria can be applied by automated scoring agents to any number of system outputs without additional clinician effort. The question is whether such rubrics can be reliably authored and validated. Importantly, we examine whether expert clinicians must author rubrics in the first place, or whether large language models (LLMs) can generate rubrics at comparable quality for a fraction of the cost. 

In this paper, we present four contributions. First, we describe and validate a case-specific, clinician-authored rubric methodology at scale: 20 clinicians authored 1,646 validated rubrics across 823 clinical cases, producing over 216,000 scored note-rubric pairs. Second, we report findings on rubric reliability, inter-clinician agreement, and scoring stability, demonstrating that the methodology discriminates effectively between high- and low-quality AI outputs and is sensitive to real system changes. Third, we present evidence that LLM-generated rubrics converge with clinician-clinician agreement as model quality improves, a finding we decompose into ``ceiling compression'' effects and genuine improvement. Fourth, we report cost analysis comparing expert and automated evaluation paths, characterizing the economics of hybrid human-LLM evaluation. These contributions are validated using Hyperscribe, an EHR-embedded clinical AI agent that converts ambient audio into structured chart updates through a five-stage pipeline \cite{hyperscribe_product,hyperscribe_github}.

\section{Materials and Methods}

\subsection{System Under Evaluation}

Hyperscribe is an EHR-embedded clinical AI agent developed by Canvas Medical \cite{hyperscribe_product}. It converts ambient clinical audio and patient context into structured chart updates \cite{hyperscribe_github}. The system processes clinical audio through a five-stage pipeline: (1) audio is transcribed into timestamped, speaker-attributed utterances; (2) the transcript is converted into clinical instructions, each representing a discrete intent mapped to a supported EHR action type; (3) instructions are translated into structured parameter objects scoped to the current encounter content; (4) parameters are converted into executable structured clinical actions (e.g., Diagnose, Prescribe, Assess, Plan, Lab Order) conforming to a typed schema and corresponding to a discrete update to the patient's chart; and (5) all interactions are logged with stage-specific labels for auditing and evaluation \cite{hyperscribe_product,hyperscribe_github,canvas_sdk}.

Each stage produces typed, schema-defined objects rather than free-form text, yielding explicit intermediate representations that can be independently inspected. This design constrains the system to a bounded set of EHR actions and makes each decision point separately evaluable--characteristics that facilitate the rubric-based evaluation methodology described in this paper. Additional architectural detail is available in a companion paper describing a holistic, end-to-end governance framework for clinical AI systems \cite{paper_b}.

\subsection{Case Definition}

We define a case as a structured representation of a clinical encounter segment. Formally, a case $C = (T, N, L)$ consists of three components: a transcript $T$ (a sequence of speaker-attributed conversational turns), a point-in-time note $N$ (the patient's documentation state at the start of the segment's audio), and longitudinal patient context $L$ (the patient's historical record including coded conditions, medications, allergies, surgical and family history).

Real-world cases ($n=736$, 168 unique patients) were extracted from live clinical encounters across primary care, psychiatry, behavioral health, oncology, and medication management settings. Transcript segments were scoped to topical exchanges within encounters. All real-world data was de-identified in accordance with HIPAA Safe Harbor requirements through a multi-pass pipeline combining automated pattern detection with manual review. Synthetic cases ($n=87$) were generated to extend coverage into clinical scenarios underrepresented in the real-world corpus, including serious mental illness, complex primary care, and medication management encounters. These cases also enabled study of the impact of synthetic provenance on expert validator behavior.

\subsection{Rubric Definition}

For each case $C$, a rubric $R$ is defined as a set of weighted criteria:

\begin{equation}
R = \{(c_i, w_i) \mid i = 1, \ldots, k\}
\end{equation}

where $c_i$ is a natural-language documentation requirement and $w_i > 0$ is a numeric weight reflecting clinical importance. Both clinician-authored and LLM-generated rubrics conform to this structure: each criterion $c_i$ is a natural-language statement prefixed with ``Reward for'' describing a specific documentation requirement, and each weight $w_i$ is an integer reflecting the criterion's clinical importance. The number of criteria $k$ ranges from 1 to 15 for clinician rubrics and is constrained to $\{4, 5, 6\}$ for LLM-generated rubrics. Two criteria are required in every rubric regardless of source: one addressing overall completeness relative to the transcript and one addressing non-repetition of information already documented in the chart.

For a given clinical note $n$ scored against rubric $R$, the normalized score is computed according to

\begin{equation}
S(n, R) = \frac{\sum_{i} w_i \cdot s_i(c_i)}{\sum_{i} w_i} \times 100
\end{equation}

where $s_i(c_i) \in [0, 1]$ represents the satisfaction of criterion $c_i$ as assessed by an LLM-based scoring agent. This yields a normalized score on a 0--100\% scale.

\begin{figure}[t]
\centering
\includegraphics[width=\textwidth]{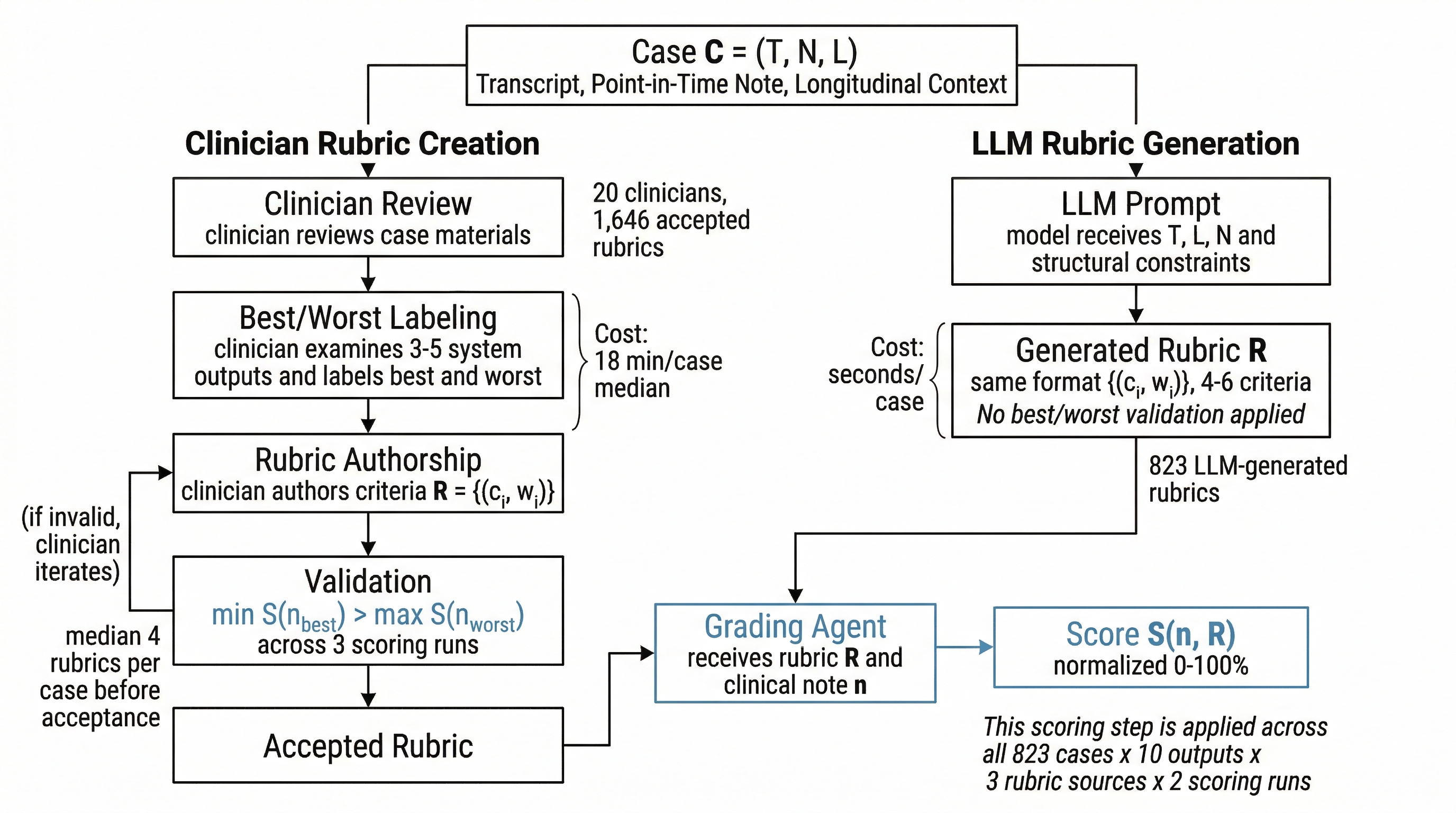}
\caption{Rubric methodology workflow. Two parallel paths for rubric creation (clinician-authored and LLM-generated) converge at a shared scoring agent. Clinician path: case review, best/worst labeling, rubric authorship, validation (min best $>$ max worst). LLM path: same case inputs, LLM prompt, generated rubric (no validation), where both are graded on the same set of cases. }
\end{figure}

\subsection{Clinician Rubric Creation and Validation}

For each case, two clinicians independently constructed rubrics (Figure 1).

The workflow proceeded as follows: (1) the clinician reviewed the transcript $T$ and longitudinal context $L$; (2) the clinician examined Hyperscribe outputs generated under different configurations alongside the point-in-time note $N$, comparing 3--5 notes produced by varying model versions and prompt strategies for the same case, and labeled the single best and single worst output based on holistic clinical judgment; (3) the clinician authored rubric criteria across dimensions including clinical accuracy, clinical action appropriateness, workflow alignment, and safety. Clinicians could begin with either an empty rubric or a starter rubric pre-populated from the case context; in both cases, the clinician retained full authority over the final content.

The best-worst labeling serves a specific purpose in validation. After the clinician authors a rubric, we test whether the rubric, when applied by the scoring agent, reproduces the clinician's ranking. If the rubric consistently scores the clinician's preferred note higher than the note the clinician judged to be worst, we can be confident that the rubric encodes the clinician's quality judgment in a form that an automated scorer can reliably execute. Formally, a rubric $R$ is accepted as valid if and only if:

\begin{equation}
\max_{j} S(n_{\text{worst}}, R, j) < \min_{j} S(n_{\text{best}}, R, j)
\end{equation}

across three independent scoring runs indexed by $j$, where $n_{\text{best}}$ and $n_{\text{worst}}$ are the clinician-identified best and worst Hyperscribe outputs for the case. This criterion ensures that even in the worst-case scoring run of the best note, the score still exceeds the best-case scoring run of the worst note, serving as a strict test of discriminative consistency.

\subsection{LLM Rubric Generation}

To examine whether the rubric creation process could be replicated without clinician involvement, we implemented an automated rubric generation pipeline. In the same way that clinicians reviewed case materials and authored criteria reflecting appropriate documentation behavior, an LLM was provided with the same case inputs ($T$, $L$, $N$) and prompted to generate a rubric $R$ conforming to the structural definition above.

The LLM was instructed to assess documentation fidelity: how accurately a note captures what was said or implied in the transcript, using relevant context from the chart. We scoped evaluation to documentation fidelity rather than clinical decision quality because Hyperscribe is a documentation tool, not a diagnostic tool. The relevant question is whether Hyperscribe is able to accurately capture the encounter and organize the information according to EHR-specific data structures. The full prompt template, JSON schema, and input field specifications are documented in a companion Github repository containing scripts and prompts for working with the dataset \cite{dataset_usage,physionet}.

The output format for LLM-generated rubrics is identical to clinician-authored rubrics: a set of $(c_i, w_i)$ pairs conforming to the same schema defined above. This structural equivalence enables direct comparison of scoring behavior across rubric sources. Unlike clinician-authored rubrics, LLM-generated rubrics were not subjected to the best-worst validation criterion; they were applied directly to the same note sets and scored using the same scoring agent, enabling comparison of ranking agreement between rubric sources.

\subsection{Scoring Methodology}
Scoring was performed by an LLM-based scoring agent prompted to evaluate each criterion $c_i$ against the clinical note and assign a satisfaction score $s_i(c_i) \in [0, 1]$. The scoring agent prompt and methodology are documented in the companion repository \cite{dataset_usage}. 

For each of seven experimental versions of Hyperscribe, 10 outputs were generated per case (5 from OpenAI models, 5 from Anthropic models) across all 823 cases. Each output was scored against three rubric sources \textemdash \space two clinician-authored rubrics and one LLM-generated rubric \textemdash \space with two scoring runs per rubric. Scores were computed using the normalized scoring functions $S(n, R)$ defined above. 

Inter-rater reliability was assessed using Kendall's tau rank correlation computed per case within each experiment, comparing: (a) clinician-clinician agreement (rankings produced by the two clinician rubrics applied to the same notes), and (b) clinician-LLM agreement (each clinician rubric paired with the LLM rubric). Score stability was assessed via the range of repeated scoring runs for each note-rubric pair. 

\subsection{Cost Tracking}
We characterized the economic cost of expert and automated evaluation through this rubric-based methodology across three dimensions. Clinician validation effort was tracked through session-level logging with evaluators reporting weekly hour totals across all validation activities (case review, provenance classification, best-worst ranking, rubric authorship, and scoring review). For LLM rubrics, inference costs were calculated based on per-call token counts (stratified by input and output token counts) and confirmed by API usage and billing records. The companion paper reports detailed cost analysis including case construction, inference cost for experimental evidence, and cost optimization strategies in the model switching and prompt minimization experimental configurations \cite{paper_b}.

\section{Results}

\subsection{Rubric Validation and Discrimination}

The benchmark dataset comprises 823 cases (736 real-world, 87 synthetic), each representing a unique patient-clinician interaction segment (Table 1). Transcripts contained a median of 419 words across 37 conversational turns, with substantial variation reflecting encounters ranging from brief follow-ups to extended multi-topic consultations (range: 28--9,402 words; 1--742 turns). Cases included varying degrees of longitudinal patient context, including demographics, point-in-time notes, and various encounter types. Cases were tagged along five dimensions: specialty (35 categories, with primary care, psychiatry, and oncology most represented), encounter type (14 categories), encounter length (short 46.7\%, medium 40.8\%, long 12.5\%, based on transcript duration), problem count (single-topic 56.7\%, multi-topic 43.3\%), and acuity (low 60.6\%, moderate 35.0\%, high 4.4\%, reflecting clinical severity). Each case has two validated rubrics from distinct clinician evaluators, yielding 1,646 rubrics with a median of 5 criteria per rubric.

\begin{table}[t]
\centering
\caption{Dataset Characteristics}
\begin{tabular}{ll}
\toprule
\textbf{Characteristic} & \textbf{Value} \\
\midrule
Total cases & 823 \\
Real-world / Synthetic & 736 (89.4\%) / 87 (10.6\%) \\
Unique patients (real-world) & 168 \\
Median transcript length & 419 words, 37 turns (range: 28--9,402 words) \\
Demographics available & 100\% \\
Point-in-time notes & 90.4\% \\
Patient age & Median 60, mean 55.8, range 9--91 \\
Patient sex & Female 65.4\%, Male 34.1\%, not specified 0.5\% \\
Specialties represented & 35 categories \\
Encounter length & Short 46.7\%, Medium 40.8\%, Long 12.5\% \\
Problem count & Single 56.7\%, Multi 43.3\% \\
Acuity & Low 60.6\%, Moderate 35.0\%, High 4.4\% \\
Accepted rubrics & 1,646 from 20 clinicians \\
Total rubrics created & 5,797 (3,060 achieving validation; 1,646 selected for use) \\
Median criteria per rubric & 5 \\
Median rubrics per case & 4 \\
\bottomrule
\end{tabular}
\end{table}

Rubric-based score discrimination was strong. Across 1,646 rubric-author pairs, the clinician-selected best note scored a mean of 68.51\% higher than the worst note on the same clinician's rubric (median 82.92\%, IQR: 43.02--95.6\%). This indicates that clinician-authored rubrics, when applied autonomously by the scoring agent, effectively capture their authors' judgments on what constitutes appropriate clinical documentation for each case. Thus, rubrics that score a note highly are encoding the same quality distinctions the clinician identified through direct comparison.

Scoring stability was high. For the 10 Hyperscribe outputs generated per case across experimental versions, repeated scoring by the scoring agent produced near-identical results. The median scoring range across repeated scoring runs was 0.00\% for clinician-authored rubrics (mean: 1.54, P95: 8.00). For LLM-authored rubrics, the median scoring range was 0.75 percentage points (mean: 2.01, P95: 8.50).

These results support the practicality of LLM-based rubric scoring as an evaluation approach. Because scoring is stable across repeated applications, observed differences between system configurations are attributable to actual differences in output quality rather than variability in the scoring process itself.

\subsection{Performance Across Experiments}

A core advantage of encoding expert judgment into reusable rubrics is that each subsequent system version can be evaluated against the full benchmark. Because rubrics are applied to every version, each experiment adds more evidence at inference cost, without requiring expert review. Seven experimental versions were evaluated, each representing iterative improvements to the Hyperscribe pipeline (Table 2, Figure 2). Each experiment produced 10 outputs per case across all 823 cases, yielding over 216,000 scored note-rubric pairs. The first four experiments (control-baseline through hierarchical-detection) showed similar distributions: median scores clustered around 83--84\%, with wide interquartile ranges (Q1: 50--58\%, Q3: 89--90\%) and standard deviations of 31--34 percentage points. A marked improvement emerged in experiments 5--7 (model-updates, model-switching, prompt-minimization), where median scores increased to approximately 95\%, the lower quartile shifted from 50--58\% to 80--81\%, and Q3 reached 100\% in all three configurations. The transition between experiments 4 and 5 represents the most significant performance discontinuity, coinciding with foundation model updates, suggesting that underlying model capability improvements can drive quality gains.

\begin{table}[H]
\centering
\caption{Experiment Performance Across Seven Configurations}
\begin{tabular}{llcccc}
\toprule
\textbf{Experiment} & \textbf{Purpose} & \textbf{Median} & \textbf{Q1} & \textbf{Q3} & \textbf{Std Dev} \\
\midrule
Control baseline & Reference point & 84.0\% & 54.2\% & 90.0\% & 33.1 \\
Next branch & Prompts, guardrails & 83.9\% & 50.7\% & 90.0\% & 33.9 \\
JSON schema & Structured parameters & 83.6\% & 50.0\% & 89.9\% & 33.8 \\
Hierarchical detection & Section-based detection & 83.3\% & 58.5\% & 89.0\% & 30.9 \\
Model updates & Latest model versions & 94.7\% & 81.0\% & 100.0\% & 27.9 \\
Model switching & Cost reduction strategy & 94.8\% & 81.0\% & 100.0\% & 28.3 \\
Prompt minimization & Token optimization & 94.3\% & 80.0\% & 100.0\% & 28.8 \\
\bottomrule
\end{tabular}
\end{table}

\begin{figure}[H]
\centering
\includegraphics[width=\textwidth]{figureA2.png}
\caption{Performance box plot showing median rubric scores across seven experimental versions. Experiments 1--4 show wide interquartile ranges (Q1 around 50--58\%) with medians at 83--84\%. Experiments 5--7 show compressed distributions with medians at approximately 95\%, Q1 around 80--81\%, and Q3 at 100\%. The marked discontinuity between experiments 4 and 5 coincides with foundation model updates.}
\end{figure}

Vendor preference analysis examined how clinicians, blinded to model provider identity, ranked outputs from different model providers within the best-worst labeling step of rubric validation. For each case, the two clinicians who authored rubrics had labeled the single best and single worst Hyperscribe output from among the 3--5 notes generated under different model configurations. Aggregating these rankings across all cases, Anthropic models achieved a net preference rate of +44.2 (72.1\% selected as best when evaluated), while OpenAI models achieved +13.6 (56.8\% selected as best). These results demonstrate that the rubric methodology is sensitive enough to detect vendor-level differences in clinical note quality as perceived by expert clinicians.

\subsection{LLM Rubric Convergence}

Given the structural equivalence between clinician-authored and LLM-generated rubrics, we examined the degree to which these rubric sources agreed when ranking the same set of Hyperscribe outputs. For each case within each experiment, the 10 outputs were ranked by each rubric source based on their normalized scores. Agreement between rankings was measured using Kendall's tau rank correlation coefficient, which quantifies the ordinal association between two rankings: given $N$ items ranked by two evaluators, tau measures the proportion of concordant versus discordant pairs, yielding a value from $-1$ (complete disagreement) to $+1$ (perfect agreement) \cite{kendall1938}. All rubrics were generated by the same LLM (o3, OpenAI). 

In experiments 1--4, clinician-clinician agreement was consistently stronger than clinician-LLM agreement: median tau ranged from 0.47 to 0.57 for clinician-clinician pairs versus 0.34 to 0.44 for clinician-LLM pairs, a consistent gap of 0.13--0.16 (Table 3). In experiments 5--7, this pattern reversed. LLM rubrics achieved agreement with individual clinicians that matched or exceeded clinician-clinician agreement: median tau was 0.42--0.46 for clinician-LLM versus 0.38--0.43 for clinician-clinician. Rank difference analysis confirmed the same pattern: in experiments 5--7, clinician-LLM rank differences (1.79--1.83 positions out of 10) were tighter than clinician-clinician differences (1.96--2.02 positions).

\begin{table}[H]
\centering
\caption{Kendall's Tau Convergence by Experiment}
\begin{tabular}{lccc}
\toprule
\textbf{Experiment} & \textbf{Clin-Clin Tau} & \textbf{Clin-LLM Tau} & \textbf{Delta (LLM -- Clin)} \\
 & \textbf{(median)} & \textbf{(median)} & \\
\midrule
1. Control baseline & 0.55 & 0.39 & $-$0.16 \\
2. Next branch & 0.51 & 0.36 & $-$0.15 \\
3. JSON schema & 0.57 & 0.44 & $-$0.13 \\
4. Hierarchical detection & 0.47 & 0.34 & $-$0.13 \\
5. Model updates & 0.38 & 0.45 & 0.07 \\
6. Model switching & 0.43 & 0.46 & 0.03 \\
7. Prompt minimization & 0.42 & 0.42 & 0.00 \\
\bottomrule
\end{tabular}

\vspace{0.5em}
\small\textit{Note:} Experiments 5--7 show clinician-LLM agreement matching or exceeding clinician-clinician agreement.
\end{table}

Two patterns are apparent in this convergence. First, ceiling compression: as note quality improved across experiments, scores clustered near the maximum. Standard deviation dropped from approximately 33\% to 28\%, separation margins compressed from 17--24 points to 8--9 points, and Q3 reached 100\%. When most notes score similarly well, distinguishing between them becomes harder for any evaluator, which mechanically reduces agreement metrics. Clinician-clinician tau declined by 0.12--0.14 points across experiments, consistent with this effect. Second, LLM rubric improvement: clinician-LLM tau increased by 0.02--0.08 points across the same period. If the convergence were entirely attributable to ceiling compression, clinician-LLM tau would have declined in parallel with clinician-clinician tau. Instead, it held steady or improved, and clinician-LLM pairs achieved tighter rank agreement than clinician-clinician pairs in experiments 5--7, a result that ceiling compression alone cannot explain.

Both factors appear to operate simultaneously: improved note quality compresses the score distribution while LLM-generated rubrics improve in their ability to rank notes consistently with clinician judgment. This suggests that LLM rubrics are most reliable as proxy experts when score distributions are compressed, which is precisely the regime where high-volume automated evaluation is most valuable for detecting subtle regressions.

\subsection{Expert Clinician Validator Experience}

Following the completion of experimental testing, we administered a 45-question mixed-methods survey to clinician annotators to understand their experiences constructing and validating rubrics.

Clinicians reported a median time of 18 minutes per case (range: 5--30 minutes), high confidence that their rubrics captured what constitutes appropriate documentation (median 4 out of 5), and trust in the rubric-based methodology for evaluating clinical AI prior to deployment (median 8 out of 10). Clinicians identified documentation of treatment plans during encounters, speaker attribution, and patient emotional content as the hardest clinical judgments to operationalize into rubric criteria. A recurring methodological challenge was achieving score separation between best and worst notes, which one reviewer noted as finding ``a distinction without meaningful difference.''

\subsection{Cost Results}

Clinician validation required 919 hours across 20 evaluators over five months, yielding 3,108 accepted rubrics at a mean effort of 17.7 minutes per accepted rubric, or approximately \$29.50 per rubric at \$100/hour. Clinicians self-reported a median of 18 minutes per case (range: 5--30 minutes); half reported stable time requirements throughout the validation period, while the other half reported decreases with experience.

LLM rubric generation via o3 produced 823 rubrics (one per case) using approximately 2.05 million input tokens and 1.23 million output tokens. At published pricing with an estimated 5x reasoning token overhead, total rubric generation cost was approximately \$14, at \$0.02 per rubric. This represents a $10^3$ magnitude cost difference: \$29.50 per clinician-authored rubric versus
  \$0.02 per LLM-generated rubric. The convergence analysis above establishes that this cost advantage does not come at the expense of ranking agreement in later experiments. Detailed cost analysis of case construction, experimental inference, and cost optimization strategies is reported in the companion paper \cite{paper_b}.

\section{Discussion}

\subsection{Prior Work}

Evaluating clinical AI documentation has relied primarily on generic instruments applied uniformly across diverse encounters. The Physician Documentation Quality Instrument and its successor (PDQI-9 and PDSQI-9, respectively) assess overall note quality through fixed Likert-scale items \cite{walker2017}. However, Walker et al. demonstrated near-zero inter-rater agreement when applying the PDQI-9 in an emergency department setting, suggesting that generic instruments may measure different constructs across clinical contexts. More recent efforts have scaled evaluation through clinician-authored rubrics: HealthBench applied expert-written criteria across a broad range of clinical scenarios for ChatGPT queries \cite{arora2025}. Our case-specific approach anchors each rubric to verifiable clinical elements derived from not only the transcript, but the patient's history, enabling granular detection of errors that generic instruments collapse into Likert-style rating scales. 

The need for granular, case-level evaluation is underscored by recent work on clinical safety. Asgari et al. found that 1.47\% of AI-generated clinical summaries contained hallucinations, of which 44\% were classified as those that could impact patient care. These were distributed across fabrication, causality errors, negation, and contextual errors, all of which are dependent on the information from the source encounter \cite{asgari2025hallucination}. Case-specific rubrics detect precisely these errors because each criterion maps to a discrete element from that encounter's transcript, chart data, and if available, the point-in-time note. 

Despite growing interest in clinical AI evaluation, the field remains fragmented. A recent scoping review of evaluation approaches for ambient clinical documentation identified only 7 studies meeting inclusion criteria, noted wide diversity of metrics across them, and found only 2 publicly available datasets \cite{benchmarking2025scoping}. Our methodology, paired with the public release of de-identified encounters and rubric criteria through PhysioNet \cite{physionet}, addresses both the methodological and data availability gaps.

\subsection{What Convergence Means for the Rubric Methodology}

The finding that LLM-generated rubrics achieve ranking agreement with individual clinicians matching or exceeding clinician-clinician agreement is a statement about ranking behavior, not clinical equivalence. Two rubrics can rank the same set of notes identically while emphasizing different clinical dimensions. What this result validates is that the rubric methodology can dramatically reduce the cost of generating evaluation evidence. At three orders of magnitude lower cost per rubric, LLM-generated rubrics enable evaluation coverage that would be economically infeasible with clinician authorship alone, accelerating the rate at which teams can iterate on AI system quality. 

This does not mean that LLMs understand clinical reasoning or that clinician involvement in rubric authorship can be eliminated. The convergence is observed in the specific task of ranking AI-generated notes by quality within a case. The clinical expertise required to define what correctness means for each case still remains a task requiring clinicians with access to the full case context. However, the proportion of LLM-generated rubrics within an evaluation system represents a practical lever for governance: a higher proportion reduces cost and increases coverage, while a sufficient proportion of clinician-authored rubrics maintains grounding in expert judgment and provides the agreement baseline against which LLM rubrics are validated. Further research is needed to characterize the optimal mix across clinical domains and system maturity levels.

\subsection{Ceiling Compression as a Methodological Caution}

An important methodological observation emerges from scrutinizing the convergence finding. As AI systems improve, their outputs cluster near maximum measurable performance, compressing the score distribution and mechanically reducing inter-rater reliability metrics. In our data, clinician-clinician tau declined from 0.47--0.57 to 0.38--0.43 as note quality improved, not because clinicians became less consistent, but because the ranking task became inherently more nuanced when most notes score similarly well.

This creates a measurement challenge relevant beyond this study: the better a system performs, the harder it becomes for any bounded evaluation measure to distinguish between its outputs. We recommend that future work reporting convergence or inter-rater agreement for clinical AI evaluation include distributional statistics (score variance, IQR, separation margins) alongside agreement metrics, enabling decomposition of ceiling effects from genuine changes in evaluator behavior. 

\subsection{The Hybrid Evaluation Model}

Rubric-based evaluation involves two roles: a rubric author who defines case-specific criteria, and a scoring agent who applies those criteria to clinical notes. Either role can be filled by a clinician or an LLM, producing four configurations:

\begin{itemize}
\item \textbf{Clinician author, clinician scorer}: highest validity but prohibitively expensive for repeated evaluation.
\item \textbf{Clinician author, LLM scorer}: preserves expert judgment in rubric design while enabling automated scoring at scale.
  This is the primary configuration validated in this study.
\item \textbf{LLM author, LLM scorer}: fully automated at approximately 1,000 times lower cost per rubric. Our convergence analysis
  demonstrates that this configuration produces ranking agreement comparable to clinician-clinician agreement in later experiments.
\item \textbf{LLM author, clinician scorer}: inefficient, as it applies expensive expert time to the lower-value scoring task rather
  than the higher-value authorship task.
\end{itemize}

The practical recommendation is not to choose one configuration but to combine them. LLM-authored, LLM-scored rubrics enable high-coverage routine evaluation and regression testing at marginal cost. Clinician-authored rubrics, also scored by LLMs, provide the expert-grounded baseline against which LLM rubric quality is continuously validated. The proportion of clinician-authored rubrics within the system is a governance parameter: sufficient clinician involvement maintains grounding in expert judgment, while the cost advantage of LLM rubrics enables evaluation coverage that would otherwise be infeasible.

The rubric methodology described here is one component of a broader evaluation and governance approach. While rubric-based evaluation captures clinical quality differences between system configurations, other evaluation dimensions, including live user feedback on workflow integration and user experience, capture failure modes invisible to static rubrics. A companion study describes the integration of these evaluation dimensions within a continuous governance framework \cite{paper_b}.

\section{Limitations}

There are limitations that constrain the generalizability of these findings. All experiments were conducted on a single AI system (Hyperscribe) within a single EHR (Canvas Medical). This methodology may require adaptation for tools that produce free-form text or operate outside an EHR.

Furthermore, our methodology is scoped to documentation fidelity and does not assess downstream care outcomes or clinical reasoning quality. However, it is our opinion that this rubric methodology is a framework to approach evaluation and governance for clinical AI tools with context-dependent expectations.

\section{Conclusion}

This study validates a rubric-based methodology for extremely cost-effective evaluation of clinical AI documentation and demonstrates that LLM-generated rubrics converge with clinician-clinician agreement for ranking AI outputs. Across 823 clinical scenarios, clinician-authored rubrics discriminated effectively between high- and low-quality documentation and detected meaningful performance differences across seven system configurations.

The convergence finding, where LLM-generated rubrics achieved ranking agreement (tau 0.42--0.46) matching or exceeding clinician-clinician agreement (tau 0.38--0.43) in later experiments, is partially attributable to ceiling compression but not entirely so. Clinician-LLM agreement held steady or improved as the ranking task became harder, indicating that automated rubrics capture quality signals aligned with clinical judgment, particularly for high-performing systems where fine-grained discrimination matters most.

The practical contribution is a hybrid evaluation model combining clinician-authored and LLM-generated rubrics, both applied by automated scoring agents. At three orders of magnitude lower cost, LLM-generated rubrics enable evaluation coverage that would be infeasible with clinician authorship alone, while continued clinician authorship provides the expert-grounded baseline against which LLM rubrics are validated. This approach maintains the connection to clinical judgment that offline benchmarks lack while operating at the cadence and cost that modern AI development and governance demands. Independent validation across multiple EHR systems, clinical domains, and AI architectures is needed to establish the generalizability of both the methodology and the convergence finding.

\section{Acknowledgments}

We thank Paulius Mui and the XPC (X Primary Care) community for organizing and collaborating with the clinicians who authored and validated the evaluation rubrics for this dataset. We thank the following clinicians for their rubric contributions accompanying the clinical scenarios: Amy Close, Fabiano Araujo, Laura Offutt, Sue StPierre, Aida Rutledge, Kevin Stephens, Elizabeth Jimenez, Ganga Nadella, Denton Shanks, Ann Pongsakul, Renee Dua, Caryn Mark, Natalie Freels, Yimdriuska Magan, Timothy Tsai, Rachel Menon, Alexandra Ristow, Casey Kirkhart, and Kevin Volkema.

We also thank the clinician users at Canvas Medical customer organizations who used Hyperscribe during its experimental phase and provided feedback that informed system improvements. 

\section{Competing Interests}

AS, AH, AD, and DB are employees or contractors of Canvas Medical and developed the Hyperscribe platform from which clinical encounter data was collected and organized. PM, FA, LO, AR, and EJ served as clinician evaluators for the rubric validation effort along with contributing to the copy of this research. Canvas Medical funded the clinician annotation effort that produced the evaluation rubrics. The dataset was prepared for public release to support independent research on clinical documentation evaluation. The authors have no financial relationships with LLM model providers beyond standard API service agreements.

\section{Funding}

This work was funded by Canvas Medical. No external funding was received.

\section{Ethics Statement}

All clinical data was de-identified in accordance with HIPAA Safe Harbor standards. Institutional review board (IRB) approval was not required.

\section{Data Availability}

The de-identified dataset is available through PhysioNet \cite{physionet}. Companion scripts for data loading, scoring, and LLM rubric generation are available in the dataset usage repository \cite{dataset_usage}. The Hyperscribe source code is publicly available \cite{hyperscribe_github}.


\end{document}